\documentclass{article}
\usepackage{ijcai26}
\usepackage{times}
\usepackage{soul}
\usepackage{url}
\usepackage[hidelinks]{hyperref}
\usepackage[utf8]{inputenc}
\usepackage[small]{caption}
\usepackage{graphicx}
\usepackage{amsmath}
\usepackage{amsthm}
\usepackage{booktabs}
\usepackage{algorithm}
\usepackage{algorithmic}
\usepackage[switch]{lineno}

\renewcommand{\thesection}{\arabic{section}.0}
\renewcommand{\thesubsection}{\thesection\expandafter\gobble\clearfield.\arabic{subsection}}
\makeatletter
\renewcommand{\thesubsection}{\arabic{section}.\arabic{subsection}}
\makeatother

\title{A Comparative Evaluation of Deep Learning Object Detection Models on a Real-World Multi-Plant Dataset from Africa}

\author{
    Ismail Ismail Tijjani$^{1,2}$\and
    Sunusi Muhammad Ibrahim$^{1,3}$\and
    Amina Ibrahim Khaleel$^{2}$\and
    Lanre Olusegun Akinola$^{4}$\and
    Fatima Isa Jibrin$^{5}$\and
    Muhammad Bashir Aliyu$^{1,6}$\and
    Abdullahi Abdussalam Dalhat$^{1,7}$\and
    Abdullahi Suiudeen$^{8}$
    \affiliations
    $^1$EJAZTECH.AI\\
    $^2$Department of Mechatronics Engineering, Bayero University, Kano, Nigeria.\\
    $^3$Department of Chemical and Petroleum Engineering, Bayero University, Kano, Nigeria.\\
    $^4$Department of Artificial Intelligence and Cybersecurity, Alpen-Adria-Universität Klagenfurt, Austria.\\
    $^5$Department of Computer Science, Gombe State University, Nigeria.\\
    $^6$Department of Mechatronics Engineering, Aliko Dangote University of Science and Technology, Wudil, Nigeria.\\
    $^7$Department of Computer Engineering, Ahmadu Bello University, Zaria, Nigeria.\\
    $^8$Department of Computer Science, Federal University Dutse, Nigeria.
    \emails
    ismailismailtj@gmail.com
}

\begin{document}

\maketitle

\begin{abstract}
The application of computer vision in agriculture has shown significant potential for improving crop monitoring and precision farming. However, many existing approaches rely on controlled datasets that do not adequately represent real-world farming conditions, particularly in underrepresented regions such as Africa. This study presents a comparative evaluation of six object detection models YOLOv5, YOLOv8, YOLO11, YOLO26, Faster R-CNN, and RT-DETR using a real-world dataset, AgriAISeg \footnote{\url{https://www.kaggle.com/datasets/ismailismailtijjani/agriaiseg}}, collected manually from Nigerian farms.
AgriAISeg comprises 3,382 images of sesame, cabbage, and tomato crops captured under varying environmental conditions, including changes in illumination, occlusion, and viewing perspectives. Models were trained, and performance was assessed using precision, recall, mAP@0.5, and mAP@0.5:0.95. The results show that RT-DETR achieved the highest overall performance with a precision of 0.768 and mAP@0.5:0.95 of 0.624, while YOLOv8 and YOLO11 also demonstrated strong and consistent performance. In contrast, Faster R-CNN recorded significantly lower accuracy, with an overall mAP@0.5 of 0.466, indicating reduced effectiveness under complex field conditions. In addition, YOLO-based models exhibited superior training efficiency compared to Faster R-CNN.These findings demonstrate that modern one-stage and transformer-based detectors provide more reliable and efficient solutions for plant detection in real-world agricultural environments.
\end{abstract}

\section{Introduction}
Agriculture serves as the primary engine for global economic stability and remains the most critical sector for ensuring food security for a growing population. Agriculture is the backbone of human civilization; however, the rising demand for food, driven by a rapidly increasing population, presents a substantial challenge to farmers and scientists who require immediate and sustainable solutions [George et al., 2025]. Furthermore, the sector faces unprecedented hurdles linked to environmental unpredictability and climate change [Delfani et al., 2024]. To mitigate these risks, there is an urgent need to integrate artificial intelligence and computer vision for sustainable crop management. Automated plant leaf disease recognition technologies have emerged to enhance the ability to efficiently identify and diagnose issues, thereby supporting improved agricultural outcomes [George et al., 2025]. Despite the potential of automated monitoring, the practical deployment of plant detection in real-world agricultural settings remains a complex technical hurdle. Images collected under real, uncontrolled illumination conditions involve varying soil backgrounds, shadows, and lighting intensities that change over several days [López-Correa et al., 2022]. In the context of smallholder farms, these difficulties are amplified by dense planting conditions and biological occlusion where leaves and stems overlap. Identifying specific species in such heterogeneous environments remains a significant challenge, as overlapping structures often lead to increased false negatives in object detection models [Beloiu et al., 2023]. 

A review of recent literature reveals critical gaps regarding the datasets used to train these systems. While models can achieve high accuracy rates, such as the 97.42 percent achieved by the RTF-RCNN model, such research often relies on isolated leaf images from repositories like PlantVillage [Alruwaili et al., 2022]. When models are trained on open-source image repositories that do not represent true field complexity, their reliability often collapses upon transition to actual field conditions due to the unpredictability of natural scenes. Furthermore, there is a critical geographical gap in available data. African agricultural landscapes and regionally specific crops are significantly underrepresented in public datasets, preventing the development of models robust enough for local ecological conditions [George et al., 2025]. In addition to data limitations, current research often lacks comprehensive benchmarking. Most existing studies limit their evaluation to only two architectures, specifically YOLOv5 and YOLOv8 [Ahmed and Abd-Elkawy, 2024]. This narrow focus fails to provide the cross-model evaluation necessary to determine which architectures are truly robust for field deployment. Without comparing one-stage, two-stage, and transformer based models on the same complex dataset, it remains difficult to select the most efficient solutions for hardware-constrained environments. To address these limitations, this research seeks to bridge the divide between theoretical performance and practical field reliability in the African context. We present a multi-model comparative evaluation across three crop classes: cabbage, tomato, and sesame.

The key contributions of this paper are:
\begin{itemize}
    \item Natural-Scene African Dataset:We introduce and publicly release AgriAISeg, a pixel-level plant image segmentation dataset for cabbage, tomato, and sesame crops collected under real-world African farming conditions. The dataset captures diverse environmental variations, including illumination changes, occlusion, and complex backgrounds.
    \item Detection Analysis under Complex Conditions: We analyze the performance of plant detection in uncontrolled environments, specifically addressing the impact of varying soil backgrounds, biological occlusion, and changing illumination on model accuracy.
    \item Extensive Cross-Model Comparison: We conduct a large-scale performance evaluation across six architectures, including the YOLO family (v5, v8, v11, and 26), Faster R-CNN, and RT-DETR, to identify the most efficient solutions for real time deployment.
    \item Open-Source Resource: To support reproducibility and future research in tropical agriculture, we provide the research community with an open-source version of our dataset and benchmarks.
\end{itemize}
The results of this study confirm that transformer-based and modern one-stage models significantly outperform traditional two-stage architectures in complex field environments. Our comparative evaluation identifies RT-DETR and the latest YOLO iterations as the most robust solutions, demonstrating superior detection accuracy and training efficiency despite biological occlusion and background noise. These findings support the deployment of lightweight, high-performance architectures for real-time agricultural monitoring in resource-constrained African farming environments.

\section{Related Work}
Artificial Intelligence (AI) has become a key driver of precision agriculture, enabling automated crop monitoring, disease detection, yield estimation, and weed management through machine learning and computer vision techniques [Olsen et al., 2019; Buzzy et al., 2020]. Deep learning models, particularly convolutional neural networks (CNNs), have significantly improved the extraction of complex visual features, thereby enhancing detection and classification performance [Buzzy et al., 2020]. Existing studies are conducted under controlled conditions, limiting their real-world applicability. Models trained on curated datasets often fail to generalize effectively in dynamic agricultural environments characterized by varying lighting conditions, occlusion, and background complexity [Wang et al., 2022]. This highlights a critical gap between experimental accuracy and field performance. Object detection has emerged as a dominant approach in plant analysis due to its ability to simultaneously localize and classify plant instances within complex scenes. Existing methods can be broadly categorized into one-stage detectors, two-stage detectors, and transformer-based models. One-stage detectors, particularly YOLO-based models, dominate due to their real-time performance and computational efficiency, making them suitable for agricultural applications such as weed detection [Beloiu et al., 2023], though these models often struggle with small, overlapping, and densely packed objects.

Two-stage detectors, such as Faster R-CNN, provide improved localization accuracy in complex environments but are computationally intensive and less suitable for real-time deployment [Chin et al., 2023; Alruwaili et al., 2022]. Transformer-based models leverage attention mechanisms for global contextual understanding but remain underexplored in agricultural applications [Babu and Venkatram, 2024]. Notably, most studies lack systematic comparisons across multiple architectures under consistent experimental conditions. Model performance is highly dependent on dataset characteristics. Controlled datasets with uniform conditions enable high accuracy but fail to capture real-world agricultural complexity [Abulizi et al., 2025]. In contrast, real-world datasets introduce variability such as illumination changes, occlusion, and background clutter, leading to reduced performance [Ahmed and Abd-Elkawy, 2024]. Most studies focus on single-crop datasets, with limited availability of multi-plant data. This limitation is particularly evident in African contexts, where publicly available datasets remain scarce [Delfani et al., 2022]. Real-world plant detection is further challenged by occlusion, lighting variability, and background complexity, all of which degrade model performance [Shehu et al., 2025; López-Correa et al., 2022; Wang et al., 2022]. Generalization remains a critical issue, as models trained on limited datasets often fail when applied to unseen environments [Saleem et al., 2025]. Additionally, computational constraints hinder deployment in resource-limited agricultural settings [Papazoglou et al., 2025].

Early work by [Olsen et al. 2019] demonstrated CNN-based weed classification using the DeepWeeds dataset but lacked real-world applicability. Subsequent studies have largely focused on YOLO-based models due to their real-time capabilities. [Buzzy et al. 2020] reported strong performance in plant-related detection tasks, although accuracy declined in dense conditions. Other researchers, including [Babu and Venkatram 2024] and [Aldakheel et al. 2024] echoed these findings, reporting high detection accuracy while highlighting persistent challenges in small object detection and generalization. Two-stage models such as Faster R-CNN improved detection accuracy in complex environments but introduced computational limitations [Alruwaili et al., 2022]. Studies integrating AI into agricultural systems consistently report reduced performance under real-world variability [López-Correa et al., 2022], while drone based approaches improve scalability but introduce noise-related challenges [Beloiu et al., 2023; Chin et al., 2023]. Many studies focus on classification rather than detection, limiting practical applicability in real-world scenarios where spatial localization is essential [Tirkey et al., 2023; Ray et al., 2025]. Although advanced YOLO variants demonstrate incremental improvements, generalization challenges persist [Ahmed and Abd-Elkawy, 2024; Abulizi et al., 2025]. Alternative approaches such as hyperspectral imaging and data-centric methods show promise but face scalability and validation limitations [Zidi et al., 2025; Papazoglou et al., 2025; Saleem et al., 2025; Joshi et al., 2025]. Overall, three key trends emerge from the literature. First, YOLO-based models dominate due to their balance between speed and accuracy, although they struggle in complex environments [Babu and Venkatram, 2024]. Second, the over-reliance on controlled datasets results in poor generalization to real-world conditions [Wang et al., 2022; Ahmed and Abd-Elkawy, 2024]. Third, many studies emphasize classification over detection, limiting real-world applicability [Tirkey et al., 2023]. The lack of diverse datasets, particularly from African contexts, remains a major limitation, alongside unresolved deployment challenges related to computational efficiency [Delfani et al., 2022; Papazoglou et al., 2025]. To address these gaps, this study conducts a comparative evaluation of multiple state-of-the-art object detection models using a locally collected real-world dataset from an African agricultural context, contributing toward more robust and deployable AI solutions in precision agriculture.

\section{Method}

\subsection{Data collection and Description}
Prior to the development of any machine learning model, the collection of a high-quality dataset is a fundamental requirement. In this study agricultural dataset was manually collected from multiple locations across Nigeria to ensure diversity in environmental conditions and crop characteristics. The dataset comprises three crop classes sesame, cabbage, and tomato, each collected from distinct geographical regions. The sesame (\textit{Sesamum indicum}) dataset was collected from Jirdede, Daura Local Government Area of Katsina State, Nigeria. A total of 891 images were captured for this class. The cabbage (\textit{Brassica oleracea}) and tomato (\textit{Solanum lycopersicum}) datasets were collected from Kura Local Government Area in Kano State, Nigeria, consisting of 1,198 and 1,293 images, respectively. All images were captured using an iPhone 11, which features a 12-megapixel dual-camera system with wide and ultra-wide lenses, capable of capturing high resolution images suitable for computer vision tasks. Data acquisition was conducted during daylight hours to leverage natural illumination conditions for real agricultural environments. To ensure robustness and improve model generalization, images were captured under varying conditions and perspectives. Specifically, the dataset includes images taken from multiple angles, including vertical (top-down), horizontal (side view), and oblique perspectives. Variations in camera to object distance were intentionally introduced, with some images captured at close range and others at moderate distances. This approach was designed to simulate real world deployment scenarios, where variations in viewpoint, scale, and background complexity are inevitable. The dataset was collected at the early growth stages of the crops, where plant structures are relatively small and more susceptible to occlusion and background interference. Figure 1 to 3 illustrates representative samples of the raw dataset for the three crop classes under varying environmental conditions.
The dataset, referred to as AgriAISeg, is publicly available for research purposes at:
\url{https://www.kaggle.com/datasets/ismailismailtijjani/agriaiseg}

\begin{figure}[ht]
\centering
\includegraphics[width=0.8\columnwidth]{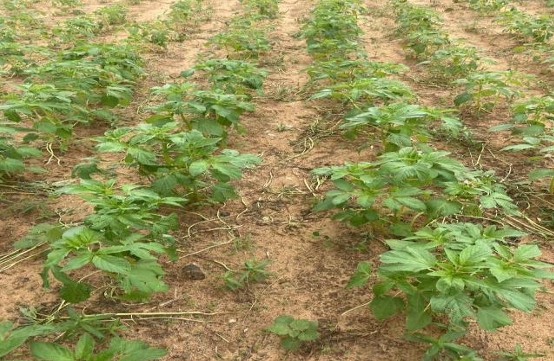}
\caption{Sesame plant Dataset}
\end{figure}

\begin{figure}[ht]
\centering
\includegraphics[width=0.8\columnwidth]{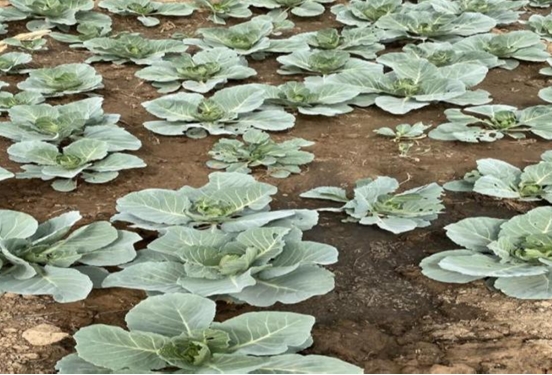}
\caption{Cabbage Plant Dataset}
\end{figure}

\begin{figure}[ht]
\centering
\includegraphics[width=0.8\columnwidth]{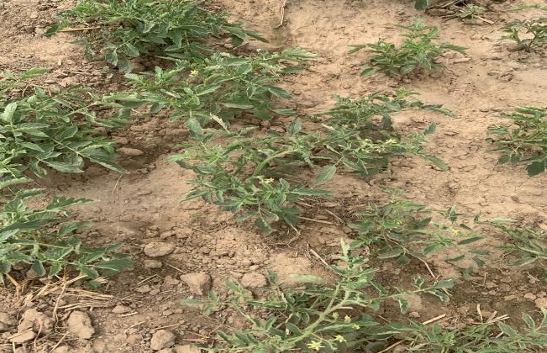}
\caption{Tomato plant Dataset}
\end{figure}

\subsection{Data Annotation}
Annotation is the fundamental stage in the development of computer vision systems, as it provides the ground truth required for model learning. It involves labeling images or marking specific regions of interest such that models can learn to accurately recognize and localize objects within complex scenes. The quality and precision of annotation influence model performance, particularly in real world environments where visual variability is high. In this study, the annotation process was carried out using the Segment Anything Model (SAM3) integrated within the Roboflow platform. Unlike conventional manual annotation approaches, which are often time-consuming and prone to inconsistency, SAM3 enables semi-automated, high precision segmentation by generating detailed masks around objects of interest. The dataset, named AgriAISeg, was originally developed as a pixel-level segmentation dataset to capture detailed plant structures under real-world conditions. However, for the purpose of this study, which focuses on object detection, the segmentation masks were systematically converted into bounding box annotations. This conversion preserves object localization while enabling compatibility with detection models. Such a transformation is widely adopted in computer vision when adapting segmentation datasets for detection tasks. The annotation process was conducted under the supervision of local farmers, who serve as domain expertise to ensure that each labeled instance accurately corresponded to the target crop. Their involvement helped maintain real-world validity and reduced the likelihood of mislabeling, especially in cases where visual similarities between crops or background vegetation could introduce ambiguity. Following the segmentation process, the annotated dataset was exported into formats compatible with the selected detection models. Although the initial labeling was performed using instance segmentation, the training process for this study was conducted using bounding box-based object detection. To facilitate this, the segmentation masks were automatically converted into bounding box annotations during export. The dataset was exported in YOLO format for training the YOLO-based models, while the COCO format was used for training the Faster R CNN and RT-DETR models.

\subsection{Detection Models}
To evaluate the effectiveness of plant detection under real-world agricultural conditions, multiple state of the art object detection models were employed in this study. The selection of models was guided by the need to compare different detection paradigms, including one-stage detectors, two-stage detectors, and transformer based approaches, to identify the most robust and efficient solution for field deployment.
\begin{itemize}
    \item \textbf{One stage detection:} The first group of models consists of the YOLO (You Only Look Once) family, including YOLOv5, YOLOv8, YOLO11, and YOLO26. These models belong to the class of one-stage detectors, which perform object localization and classification in a single forward pass. Their architecture is optimized for real-time performance, making them highly suitable for applications such as precision agriculture where low latency is essential. YOLO based models are known for their balance between speed and accuracy, as well as their ability to generalize effectively across varying environmental conditions.
    \item \textbf{Two stage detection:} Faster R-CNN was employed as two-stage detection model. This architecture operates by first generating regional proposals using a Region Proposal Network (RPN), followed by classification and bounding box refinement. While Faster R-CNN is widely recognized for its high localization accuracy, particularly in controlled environments, it is computationally intensive and often exhibits slower inference and training times. Its inclusion in this study provides a baseline for evaluating the trade-off between accuracy and computational efficiency.
    \item \textbf{Transformer-based detection:} RT-DETR (Real-Time Detection Transformer), was included to explore the impact of attention mechanisms in complex agricultural scenes. Unlike traditional convolutional architectures, RT-DETR leverages global context through self-attention, enabling it to better capture relationships between objects within an image. This characteristic has advantageous in scenarios involving occlusion and dense plant arrangements, where local feature extraction alone may be insufficient.
\end{itemize}
All models were trained and evaluated under consistent experimental conditions using the same dataset and preprocessing pipeline. This ensures a fair and unbiased comparison, allowing for a comprehensive assessment of each model’s performance in terms of detection accuracy, robustness to environmental variability, and computational efficiency.

\subsection{Training Configuration and Experimental Setup}
All models were trained under a unified experimental framework to ensure a fair and consistent comparison across different detection models. The dataset was divided into 75\% training, 15\% validation, and 10\% testing, providing sufficient data for model learning, hyperparameter tuning, and unbiased evaluation on unseen samples. All input images were resized to a fixed resolution of (640 × 640) to maintain consistency across all models and ensure compatibility with the respective detection frameworks. For training configurations, Faster R-CNN was explicitly tuned due to its sensitivity to hyperparameter selection, while YOLO-based models and RT-DETR were trained using their recommended default settings. The hyperparameter settings for each model are summarized in Table 1.

\begin{table}[ht]
\centering
\resizebox{\columnwidth}{!}{
\begin{tabular}{@{}llllll@{}}
\toprule
Model & Learning Rate & Momentum & Batch Size & Epochs & Weight Decay \\ \midrule
Faster R-CNN & 0.005 & 0.9 & 4 & 50 & 0.0005 \\
YOLOv5 & 0.01 & 0.937 & 16 & 50 & 0.0005 \\
YOLOv8 & 0.01 (auto) & 0.937 & 16 & 50 & 0.0005 \\
YOLO11 & 0.01 (auto) & 0.937 & 16 & 50 & 0.0005 \\
YOLO26 & 0.01 & 0.98 & 16 & 50 & 0.0005 \\
RT-DETR & 0.0001 & 0.9 & 16 & 50 & 0.0001 \\ \bottomrule
\end{tabular}
}
\caption{Training Hyperparameters for Evaluated Models}
\end{table}

\begin{figure}[ht]
\centering
\includegraphics[width=0.8\columnwidth]{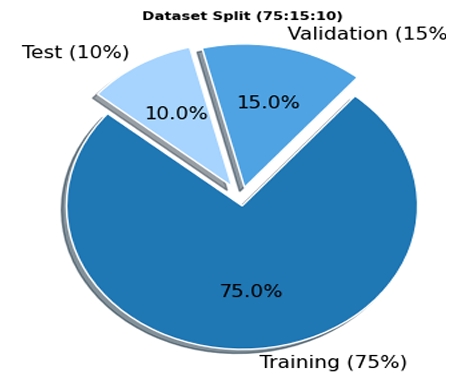}
\caption{The Dataset split into train, validation and test}
\end{figure}

\begin{figure}[ht]
\centering
\includegraphics[width=0.8\columnwidth]{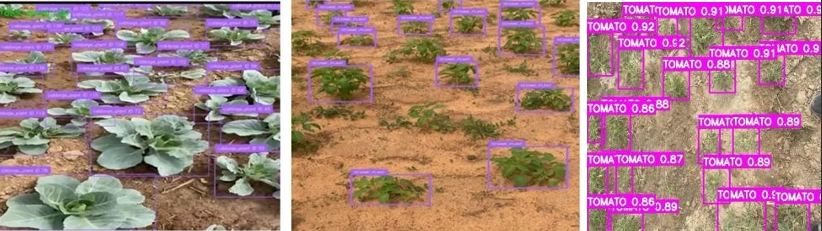}
\caption{Predicted bounding boxes for Cabbage, Sesame and Tomato plants}
\end{figure}

\subsection{Evaluation Metrics}
To evaluate the performance of the detection models, standard object detection metrics were employed. These metrics provide complementary insights into the ability of the models to accurately identify and localize plant instances within complex agricultural scenes.

\textbf{Precision (P):} measures the proportion of predicted detections that are correctly identified. It reflects the model’s ability to minimize false positive predictions.
\begin{equation}
Precision = \frac{TP}{TP + FP}
\end{equation}
where $TP$ represents true positives and $FP$ represents false positives.

\textbf{Recall (R):} measures the proportion of actual objects that are correctly detected by the model. It indicates how well the model captures all relevant instances within an image.
\begin{equation}
Recall = \frac{TP}{TP + FN}
\end{equation}
where $FN$ denotes false negatives.

\textbf{Intersection over Union (IoU):} is used to quantify the overlap between predicted bounding boxes and ground truth annotations. It is defined as the ratio between the area of overlap and the area of union.
\begin{equation}
IoU = \frac{Area \ of \ Overlap}{Area \ of \ Union}
\end{equation}

\textbf{Mean Average Precision (mAP):} is a comprehensive metric used to evaluate overall detection performance across all classes. It is computed as the Average Precision (AP) values for each class:
\begin{equation}
mAP = \frac{1}{N} \sum_{i=1}^{N} AP_i
\end{equation}
where $N$ is the number of classes and $AP_i$ represents the average precision for class $i$. Performance is reported using mAP@0.5, as well as mAP@0.5:0.95.

\textbf{F1 Score:} represents the harmonic means of precision and recall, providing a balanced measure of detection performance.
\begin{equation}
F1 = 2 \times \frac{Precision \times Recall}{Precision + Recall}
\end{equation}
These metrics provide a comprehensive and reliable evaluation of model performance, capturing both localization accuracy and detection completeness under real-world agricultural conditions.

\section{Result and Discussion}

\subsection{Quantitative Performance Analysis}
The empirical evaluation of the six object detection models revealed significant variations in performance across the Sesame, Cabbage, and Tomato classes. As summarized in Table 2, the transformer-based RT-DETR and the modern one-stage YOLOv8 and YOLO11 models consistently outperformed the traditional two-stage Faster R-CNN framework. The predicted bounding box for the all three plants are shown in Figure 5.

\begin{table}[ht]
\centering
\resizebox{\columnwidth}{!}{
\begin{tabular}{@{}llllll@{}}
\toprule
Model & Precision & Recall & mAP@0.5 & mAP@0.5:0.95 & Time (hrs) \\ \midrule
RT-DETR & 0.768 & 0.779 & 0.806 & 0.624 & 4.12 \\
YOLOv8 & 0.761 & 0.776 & 0.808 & 0.621 & 2.00 \\
YOLO11 & 0.763 & 0.772 & 0.806 & 0.618 & 2.00 \\
YOLOv5 & 0.756 & 0.774 & 0.802 & 0.611 & 1.69 \\
YOLO26 & 0.747 & 0.755 & 0.790 & 0.606 & 2.00 \\
Faster R-CNN & 0.447 & 0.345 & 0.466 & 0.240 & 8.75 \\ \bottomrule
\end{tabular}
}
\caption{Comparative Performance Summary Across All Evaluated Models}
\end{table}

\subsection{Architectural Efficacy and the Attention Mechanism}
The performance of RT-DETR (mAP@0.5:0.95 of 0.624) underscores the advantage of the Vision Transformer (ViT) backbone in complex agricultural scenes. Unlike the local receptive fields of traditional CNNs, the self-attention mechanism in RT-DETR allows the model to capture global contextual dependencies. This is particularly vital in the early growth stages where plant structures are small and highly like background weeds. While YOLOv8 achieved a slightly higher mAP@0.5 (0.808), it fell behind RT-DETR in the more rigorous mAP@0.5:0.95 metric. This suggests that while YOLO models are excellent at general localization, the transformer-based approach provides superior bounding-box refinement and spatial precision under the biological occlusion conditions characteristic of the Nigerian in form field.

\subsection{The Failure of Two Stage Detectors in Unstructured Fields}
An important finding of this study is the significant performance collapse of Faster R-CNN, which recorded a precision of only 0.447 and a recall of 0.345. Despite its reputation for high accuracy in controlled environments the model struggled with the stochastic noise and varying illumination of the African farm environment. The intensive training time of Faster R-CNN (8.75 hours) more than double that of any other model did not translate into better feature extraction. This failure is likely attributed to the Region Proposal Network (RPN) being overwhelmed by the high frequency of false proposals generated by complex soil backgrounds and overlapping leaves. In contrast, the one-stage YOLO family maintained a robust balance between computational efficiency and detection completeness.

\subsection{Per Class Performance and Environmental Sensitivity}
When analyzing individual crops, Cabbage consistently yielded the highest detection metrics across all models (e.g., YOLO11 mAP@0.5 of 0.958). This can be attributed to the relatively distinct, broad-leaf geometry of the cabbage plant, which provides high contrast against the soil. Conversely, Sesame proved to be the most challenging class (RT-DETR mAP@0.5 of 0.700). The narrow, fine structure of sesame seedlings at their early growth stage makes them highly susceptible to background interference and scale variations. These results highlight the need for regionally specific datasets models trained on generic data would likely fail to detect these specialized crop structures in a real world Nigerian agricultural deployment.

\subsection{Real Time Deployment Suitability}
For practical application in resource constrained environments, YOLOv8 and YOLOv5 emerge as the most viable candidates. Their ability to converge within approximately 2 hours while maintaining mAP@0.5 scores above 0.80 suggests a high degree of efficiency. While RT-DETR offers the highest precision, its 4.12-hour training time and greater computational requirements must be weighed against the hardware limitations of mobile devices or low-cost drones used by small-holder farmers.

\section{Conclusion}
This study presented a comprehensive evaluation of state-of-the-art object detection models for plant detection using a locally curated, real-world dataset, AgriAISeg, collected from agricultural fields in Africa. Unlike many existing works that rely on controlled or publicly available datasets, this research emphasizes the importance of using region-specific data that captures the complexity and variability of real farming environments, including occlusion, illumination changes, and diverse viewing conditions.

The comparative analysis across six detection models demonstrated that modern one-stage and transformer-based approaches are more effective for real-world agricultural applications. In particular, RT-DETR achieved the highest overall performance, while YOLOv8 and YOLO11 provided a strong balance between accuracy and computational efficiency. In contrast, Faster R-CNN showed significantly lower performance, highlighting its limitations in handling complex field conditions. Beyond model comparison, the findings underscore the important role of locally representative datasets in developing robust and deployable agricultural AI systems. Models evaluated on real-world data provide more reliable insights into practical performance than those trained solely on controlled datasets.

Furthermore, the public release of AgriAISeg provides a valuable benchmark for future research in real-world agricultural computer vision, particularly in underrepresented regions such as Africa.

\end{document}